%% file: main.tex
\documentclass[10pt,twocolumn,letterpaper]{article}

\usepackage{cvpr}              
\usepackage{float}
\usepackage{graphicx}
\usepackage{tikz}
\usepackage{tabularx}
\usepackage{array}
\usepackage{adjustbox}
\usepackage{arydshln}
\input{math_commands}

\usepackage{multirow}

\definecolor{cvprblue}{rgb}{0.21,0.49,0.74}
\usepackage[pagebackref,breaklinks,colorlinks,citecolor=cvprblue]{hyperref}

\def\paperID{16346} 
\def\confName{CVPR}
\def\confYear{2024}

\input{0_macros}

\title{\methodname{}: In-Context, Interleaved, and Interactive Any-to-Any Generation
}

\author{%
  Zineng Tang$^{1,4}\thanks{Work done while at Microsoft internship and UNC.}$\;\;\;\vspace{15pt}
  Ziyi Yang$^{2}\thanks{Correspondence: ziyiyang@microsoft.com, mbansal@cs.unc.edu}$\;\;\;
  Mahmoud Khademi$^{3}$\;\;\;
  Yang Liu$^{2}$\;\;\;
  Chenguang Zhu$^{3}\thanks{Work done while at Microsoft.}$\;\;\;
  Mohit Bansal$^{4\dagger}$\;\;\; 
  \\
  \vspace{12pt}$^1$UC Berkeley \;\; $^2$Microsoft Azure AI\;\; $^3$Zoom \;\; $^4$UNC Chapel Hill
  \\
{\textbf{\tt \url{https://codi-2.github.io}}}
}

\begin{document}

\maketitle

\input{0_abstract}

\input{1_intro}
\input{11_relatedwork}
\input{2_method}
\input{3_experiment}

\input{4_conclusion}
{
    \small
    \bibliographystyle{ieeenat_fullname}
    \bibliography{main}
}

\input{X_suppl}

\end{document}

%% file: math_commands.tex
\usepackage{amsmath,amsfonts,bm}

\def\eqref#1{equation~\ref{#1}}

\def\1{\bm{1}}

\def\vc{{\bm{c}}}

\def\vx{{\bm{x}}}
\def\vy{{\bm{y}}}
\def\vz{{\bm{z}}}
\def\veps{{\bm{\epsilon}}}

\DeclareMathAlphabet{\mathsfit}{\encodingdefault}{\sfdefault}{m}{sl}
\SetMathAlphabet{\mathsfit}{bold}{\encodingdefault}{\sfdefault}{bx}{n}

\def\gL{{\mathcal{L}}}

\newcommand{\E}{\mathbb{E}}



%% file: 0_macros.tex
\newcommand{\encoder}[1]{$E$}
\newcommand{\decoder}[1]{$D$}
\newcommand{\maskoutputv}[1]{$\hat{x}_v$}

\newcommand{\methodname}[1]{CoDi-2}

%% file: 0_abstract.tex
\begin{abstract}
We present \methodname{}, a versatile and interactive Multimodal Large Language Model (MLLM) that can follow complex multimodal interleaved instructions, conduct in-context learning (ICL), reason, chat, edit, etc., in an any-to-any input-output modality paradigm. By aligning modalities with language for both encoding and generation, \methodname{} empowers Large Language Models (LLMs) to not only understand complex modality-interleaved instructions and in-context examples, but also autoregressively generate grounded and coherent multimodal outputs in the continuous feature space. 
To train \methodname{},  we build a large-scale generation dataset encompassing in-context multimodal instructions across text, vision, and audio. \methodname{} demonstrates a wide range of zero-shot capabilities for multimodal generation, such as in-context learning, reasoning, and compositionality of any-to-any modality generation through multi-round interactive conversation. \methodname{} surpasses 
previous domain-specific models on tasks such as subject-driven image generation, vision transformation, and audio editing. \methodname{} signifies a substantial breakthrough in developing a comprehensive multimodal foundation model adept at interpreting in-context language-vision-audio interleaved instructions and producing multimodal outputs.
\end{abstract}

\begin{figure*}[t!]
  \centering
   \includegraphics[width=0.99\textwidth]{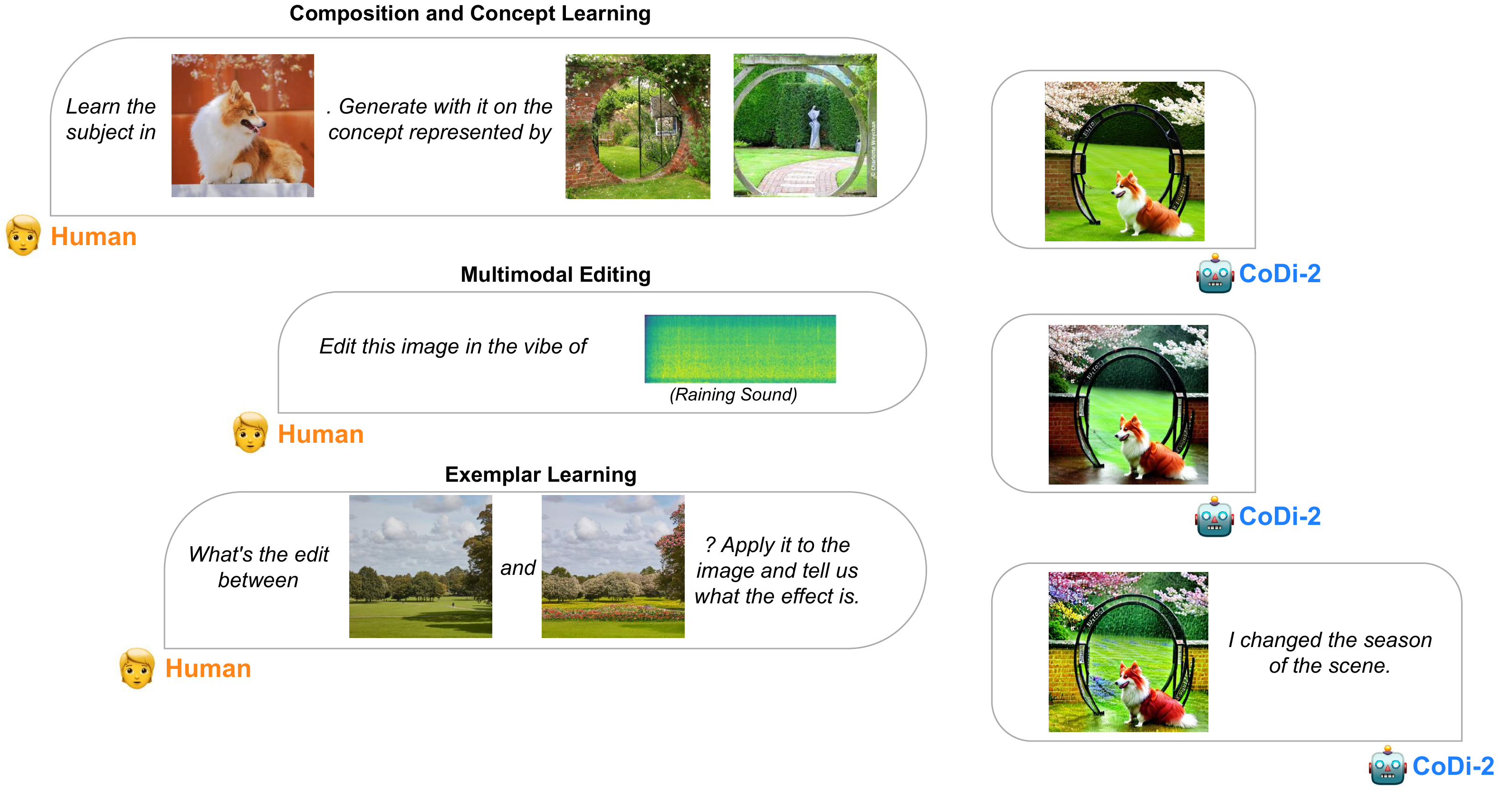}
   \caption{Multi-round conversation between humans and \methodname{} offering in-context multimodal instructions for image editing.
   }
   \label{fig:teaser}
\vspace{-10px}
\end{figure*}

%% file: 1_intro.tex
\section{Introduction}
\label{sec:intro}
Multimodal generation has achieved remarkable advancements in recent years, e.g., generating high-fidelity image, video, audio and music samples from prompt provided by users. Recent advancements in AI-Generated Content (AIGC) highlight in-context generation~\cite{pan2023kosmos,wang2023context}, concept learning~\cite{ruiz2023dreambooth}, editing~\cite{brooks2023instructpix2pix}, and fine-grained control~\cite{zhang2023adding}. Recently, \citet{tang2023any} proposed CoDi, the first model ever that can generate any combinations of modalities from any combinations of input ones. Building upon this foundational work, the subsequent study by~\cite{wu2023next} further advances CoDi by proposing a model that can facilitates conversational abilities and expansion to additional modalities.

Although remarkable advances have been made in multimodal generation, several critical challenges remain: (1) Zero-shot fine-grained and sophisticated user-control of multimodal generation is infeasible: current multimodal generative models (MGM) cannot generate sophisticated in-context generation examples without finetuning on subtasks, such as replicating or transferring an editing effect via an `analogy' setting or subject driven generation, as demonstrated in the prompt (as in the row ``Exemplar Learning'' and ``'Subject Driven' of \Cref{tab:task_types_mm}). Moreover, the reasoning ability of MGM is rather limited, e.g., the input prompts are usually descriptive where the generation do not require capabilities such as logical, compositional, and analytical intelligence.
(2) The user-and-model interaction is usually constrained to single-round, or it is challenging for current models to follow multi-round instructions while ensuring the consistency and faithfulness of responses across the rounds, as shown in \Cref{fig:teaser}. 
(3) The inputs in previous MGMs mostly only contain one or two modalities. The ability to understand modality-interleaved inputs, such as language instruction mixing with contextual visual and auditory inputs is critical to building a fundamental multimodality model. Hence, overall, a versatile any-to-any MGM, that can follow interleaved in-context multimodal instructions and interactive multi-round chatting is strongly needed.

To this end, we propose \methodname{}, a versatile Multimodal Large Language Model (MLLM) that can conduct any-to-any generation, in-context and modality-interleaved instruction following, and multi-round multimodal chat to achieve editing, reasoning, and compositionality tasks, etc. Enabling in-context learning and following interleaved multimodal instructions in multimodal generation is challenging. In previous multimodal generative models, the backbone is mostly diffusion models (DMs) which are good at generation but intrinsically lack the capability to perform in-context understanding~\cite{west2023generative}. We therefore propose to harness a Large Language Model (LLM) as the ``brain'' to understand modality-interleaved human instructions, execute in-context learning, and integrate multimodal input signals, because LLMs have strong language reasoning capabilities for complex instructions in the language domain. By mapping all modalities to the space of language (as proposed in CoDi \citep{tang2023any}) and connecting these modalities to LLM through encoder and synchronized decoders, \methodname{} can process multimodal inputs within their context by aligning image or audio features to the language model input space, and understand the delicate modality-interleaved instructions for zero-shot or few-shot generation. In this way, our system can harness and inherent the in-context learning (ICL), reasoning, chatting, zero-shot learning, instruction following capabilities of large language models (LLMs) by processing and interpreting multimodal inputs within their context by aligning multimodal (e.g., vision and audio) features to both input and output space. For generation, we propose to train the MLLM to autogressively predict the features of the output modality. The predicted features are then input to (synchronized) diffusion models. This end-to-end any-to-any generative framework enables \methodname{}
to conduct elaborate reasoning for understanding and generate multiple modalities, and therefore allows diverse tasks such imitation, editing, compositional creation, etc. In training \methodname{}, the gradient obtained from the diffusion models' generation loss also directly back-propagates to the LLM which can enhance the perceptual faithfulness to the inputs including images or audio.

The development of the alignment data to train such model is also challenging, hindered by the scarcity of specialized data such as multimodal reasoning or in-context learning. To start with, we comprehensively collect latest instructional generation datasets across vision, audio, and language. We then propose to convert these instructional datasets to in-context generation ones, such that in the prompt (e.g., row ``Exemplar Learning'' of \Cref{tab:task_types_image}) and more can be referred in \Cref{tab:task_types_image,tab:task_types_audio,tab:task_types_mm}.
To further diversify the in-context learning datasets, we propose a novel method to build text-only datasets for multimodal in-context learning. Since language and other modalities (vision and audio) are mapped to the same space through the aligned encoders, we can flexibly build multimodal datasets with only language, where the multimodal components are represented by their respective textual descriptions (e.g., using image caption instead of the pixels to represent the image).

Empirical assessments of our multimodal generation tasks, which include a diverse array of complex and intertwined instructions, yield remarkable results. These tasks encompass audio fusion and editing, image generation with intricate composition, the use of in-context exemplars, and sophisticated reasoning, as well as understanding and generating videos. This wide range of tasks show strong capability in both zero-shot and few-shot prompting settings, showcasing our system's adaptability and robust performance across different scenarios. Overall, \methodname{} stands out in integrating in-context learning within the domain of interleaved and interactive multimodal any-to-any generation. \methodname{} marks a significant step in the field of multimodal generation, achieving groundbreaking in-context, interleaved and interactive any-to-any generation.

%% file: 11_relatedwork.tex
\section{Related Work}

\subsection{Multimodal Large Language Models}
Recent years have witnessed the rapid evolution of LLMs, setting a new precedent in natural language understanding and generation~\cite{openai2023gpt4, touvron2023llama, touvron2023llama2}. Multimodal LLMs extend LLMs to multimodal learning \cite{zellers2022merlot,yang2023code}, enabling the processing of diverse input forms, not just limited to text but also incorporating visual and other sensory data~\cite{liu2023visual,you2023ferret,lu2023chameleon,ye2023mplug,gao2023llama,li2023otter}. The innovation in this space has led to models that are not only capable of understanding multimodal inputs but also adept at generating multifaceted outputs, thereby pushing the boundaries of creative and contextual AI-generated content~\cite{tang2023any, wu2023next}. Another notable line of work is using LLM to ground image generation~\cite{koh2023generating, sun2023generative}. 

\subsection{Multimodal In-Context}
Multimodal in-context requires sometimes interleaved in-context understanding of multimodal inputs like images and text like Wikipedia (with images), documents, videos with narrations or QAs, etc.
This domain has expanded yet facing its set of challenges. While there is a plethora of research focusing on the understanding aspect of multimodal data~\cite{li2023otter,zheng2023minigpt}, the generation of raw sensory perceptions such as images or audio remains a complex hurdle. The concept of treating images as a foreign language opened new avenues, particularly in in-context image generation~\cite{pan2023kosmosg}. However, these pioneering techniques are still in nascent stages, often constrained by their training regimes and lacking genuine in-context learning capabilities, which limits their performance and adaptability.

\subsection{Multimodal Generation}
Recent years have witnessed a significant growth in image editing and manipulation research, which can be broken into image editing~\cite{meng2021sdedit,brooks2023instructpix2pix}, exemplar learning~\cite{wang2023context} for image generation, image composition~\cite{ruiz2023dreambooth,kumari2023multi,pan2023kosmosg}, and concept learning~\cite{huang2023reversion} from images. 

Image editing~\cite{meng2021sdedit} uses guidance control and edit the attributes of an image. To align the guidance with human instructions, InstructPix2Pix~\cite{brooks2023instructpix2pix} takes in instructional image editing prompts to directly transform an image. The realm of image composition are tasks that compose one or more images into a single image and demand high fidelity to input images, which poses unique challenges. Techniques involved in subject-driven image generation~\cite{ruiz2023dreambooth} have shown promise in transforming a subject into a new scene. However, they often necessitate task-specific or subject-specific tuning. This specialization often confines the models within the boundaries of their training data, impeding their ability to generalize beyond learned tasks or subjects. Kosmos-G~\cite{pan2023kosmosg} furthers the efforts for zero-shot image generation with in-context interleaved image and text. But its efforts is limited to image composition. Lastly, learning visual concepts and apply them in image generation is also a growing direction~\cite{kumari2023multi,huang2023reversion}. For example, multi-concept customization to text-to-image generation~\cite{kumari2023multi} requires the model to extract visual concept like a moon gate or a certain subject and apply them in image generation. The aspiration to develop a model with in-context multimodal reasoning abilities to transcend these limitations inspires our versatile framework that takes in task instructions and perform in-context zero-shot generation.

%% file: 2_method.tex
\section{Model Architecture}

\methodname{} is designed to process in-context multimodal inputs, including text, images, and audio, utilizing specific instructions to facilitate in-context learning and generate corresponding text, images, or audio outputs. The model is distinguished by the several key features as introduced in following subsections.

\begin{figure}[t]
  \centering
   \includegraphics[width=0.5\textwidth]{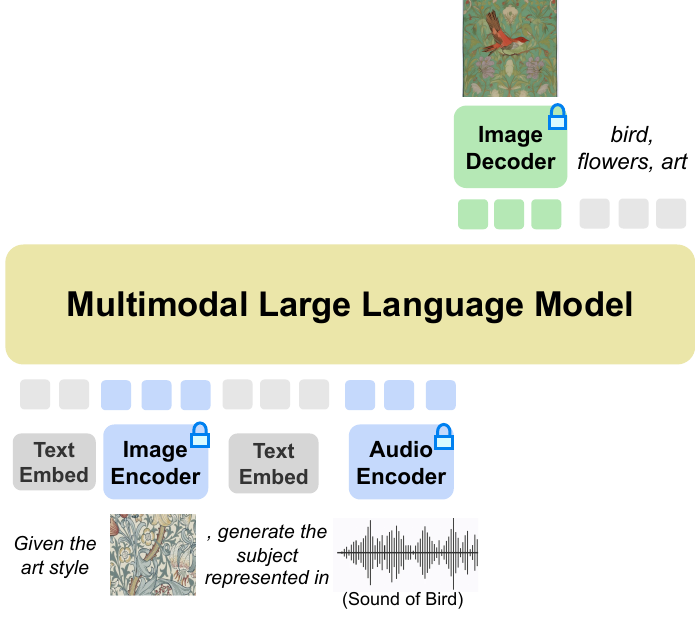}
   \caption{
   Model Architecture: \methodname{} comprises a multimodal large language model that encompasses encoder and decoder for both audio and vision inputs, as well as a large language model. This architecture facilitates the decoding of image or audio inputs using diffusion models. In the training phase, our approach employs pixel loss obtained from the diffusion models alongside token loss, adhering to the standard causal generation loss.}
   \label{fig:model}
\vspace{-10px}
\end{figure}
\subsection{Multimodal LLM as the Fundamental Engine}
Building such an any-to-any foundation model that can digest interleaved inputs of modalities, understand and reason over complex instructions (e.g., multi-round conversation, in-context examples), and interact with multimodal diffusers requires a powerful fundamental ``engine''. We propose to leverage MLLM for this engine, which is built by empowering a text-only LLM with multimodal perceptions.

The motivation of harnessing LLM is intuitively inspired by the observation that LLMs exhibit exceptional ability such as chatting, zero-shot learning, instruction-following, etc, in language-only domain ~\cite{zhao2023survey}. By leveraging projections from aligned multimodal encoders (e.g., \citep{tang2023any}), we can seamlessly empower the LLM to perceive modality-interleaved input sequence. Specifically, in processing the multimodal input sequence, we first use the multimodal encoder to project the multimodal data into a feature sequence. Special tokens are prepended and appended to the features sequence, e.g. ``\texttt{$\langle$audio$\rangle$} \texttt{[audio feature sequence]} \texttt{$\langle$/audio$\rangle$}". By such for instance, a modality-interleaved input sequence ``\textit{A cat sitting on} [\texttt{image0:an image of a couch}] \textit{is making the sound of} [\texttt{audio0:audio of cat purring}]'' is then transformed to ``\textit{A cat sitting on} \texttt{$\langle$image$\rangle$} \texttt{[image feature sequence]} \texttt{$\langle$/image$\rangle$} \textit{is making the sound of} \texttt{$\langle$audio$\rangle$} \texttt{[audio feature sequence]} \texttt{$\langle$/audio$\rangle$}, before inputting to the MLLM to process and generation.

\subsection{Multimodal Generation with MLLM}

To generate text, the MLLM can naturally generate text tokens autoregressively; for multimodal generation, one common way in previous works was to transform the multimodal target (e.g., the ground-truth image) into discrete tokens such that they can be generated autoregressively like text. However, the generation quality of this methodology is intrinsically constrained by the VAE-like generation decoder, while current SOTA multimodal generation frameworks generally adopt Diffusion Models (DMs)~\citep{rombach2022high}. Therefore, we propose to integrate DMs into MLLM to generate multimodal outputs, following nuanced modality-interleaved instructions and prompts. Recall the training objective of a diffusion model is given as:
\begin{equation}
\label{eq:df}
    \gL_{DM} = \E_{\vz, \veps, t}\|\veps - \veps_{\theta}\big(\vz_t, t, C_y(\vy)\big)\|_2^2,
\end{equation}
where $\vy$ is the conditional modality, $C_y$ is the conditional encoder for $\vy$, $\veps_{\theta}$ is the U-Net, and $\vz_t$ is the noisy latent variable at time step $t$.

We propose to train the MLLM to generate the conditional feature $\vc = C_y(\vy)$ that will be fed into DM to synthesize the target output $\vx$. By such, the generative loss of DM can be used to train MLLM. To further provide a stronger and directer supervision signal for MLLM, and to retain the perceptual characteristics inherent in the original input, we explicitly induce that $\vc = C_x(\vx)$, i.e., MLLM is trained to generate $C_x(\vx)$ and DM is expected to function as an autoencoder in this case\footnote{We leverage the condition encoder aligned across modalities from CoDi (\citet{tang2023any}, named prompt encoder in the original paper)}. The mean squared error between MLLM output feature $\vc_{MLLM}$ and $C_x(\vx)$, together with $\gL_{DM}$, and text token prediction loss $\gL_{t}$ is the final training loss: $\gL = \alpha\text{MSE}\big(\vc_{MLLM}, C_x(\vx)\big)+\gL_{DM}$ + $\gL_{t}$ controlled by weighting $\alpha$. 

%% file: 3_experiment.tex
\section{Building Diverse Multimodal In-Context Generation Data}

\newcommand\xw{0.19}
\begin{table*}[t]
\centering
\begin{tabularx}{0.999\textwidth}{m{2cm} m{0.65\linewidth}: m{3.5cm}}
\toprule
\textbf{Task Type} & \textbf{Example Prompt} & \textbf{Output}\\
\midrule
\midrule
&\textbf{Zero-Shot Prompting}& \\
\midrule
\textbf{Instruction Editing} & Turn this \adjustbox{valign=m}{\includegraphics[width=\xw\linewidth]{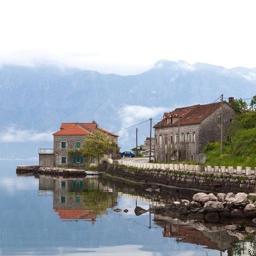}} into Van Gogh style & \adjustbox{valign=m}{\includegraphics[width=0.7\linewidth]{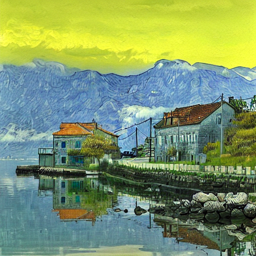}} \\
\hdashline
\textbf{Composition} & A \adjustbox{valign=m}{\includegraphics[width=\xw\linewidth]{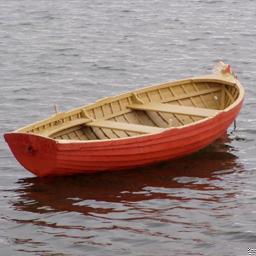}} on \adjustbox{valign=m}{\includegraphics[width=\xw\linewidth]{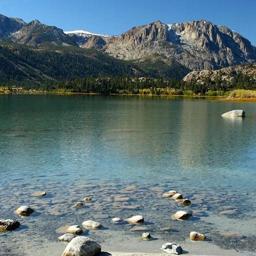}} . &  \adjustbox{valign=m}{\includegraphics[width=0.7\linewidth]{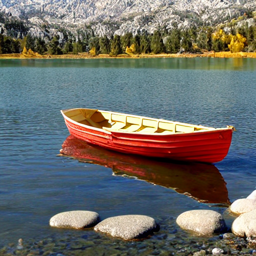}}\\
\hdashline
\textbf{Reasoning} & Given \adjustbox{valign=m}{\includegraphics[width=\xw\linewidth]{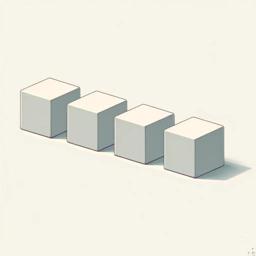}} \adjustbox{valign=m}{\includegraphics[width=\xw\linewidth]{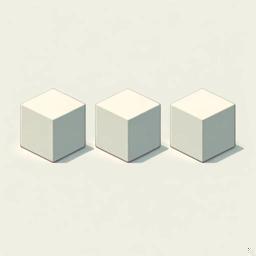}} and \adjustbox{valign=m}{\includegraphics[width=\xw\linewidth]{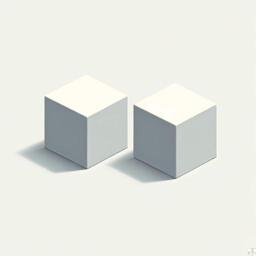}} , what happens next? & \adjustbox{valign=m}{\includegraphics[width=0.7\linewidth]{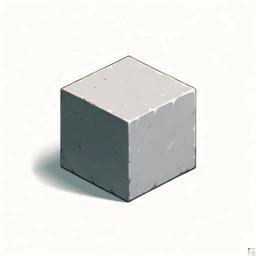}} \\
\midrule
&\textbf{One-Shot/Few-Shot Prompting}& \\
\midrule
\textbf{Exemplar Learning} & We apply a new concept to \adjustbox{valign=m}{\includegraphics[width=\xw\linewidth]{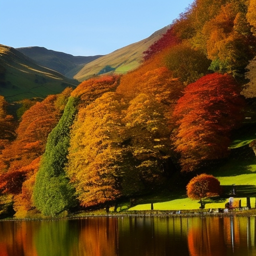}} and got \adjustbox{valign=m}{\includegraphics[width=\xw\linewidth]{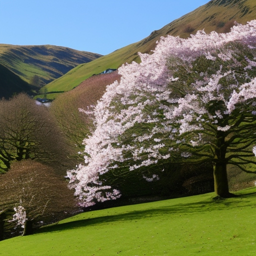}}. 

Apply the same concept to \adjustbox{valign=m}{\includegraphics[width=\xw\linewidth]{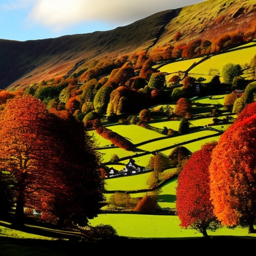}}. & \adjustbox{valign=m}{\includegraphics[width=0.7\linewidth]{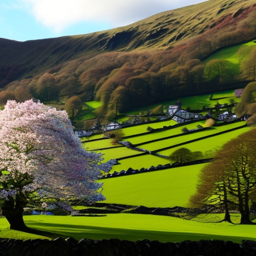}} \\
\hdashline 
\textbf{Concept Learning} & Given the artistic style represented in \adjustbox{valign=m}{\includegraphics[width=\xw\linewidth]{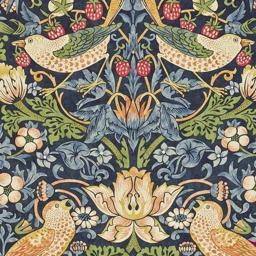}} and \adjustbox{valign=m}{\includegraphics[width=\xw\linewidth]{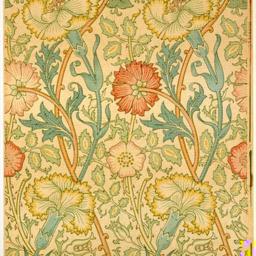}} , 

create a new artwork similar to it. & \adjustbox{valign=m}{\includegraphics[width=0.7\linewidth]{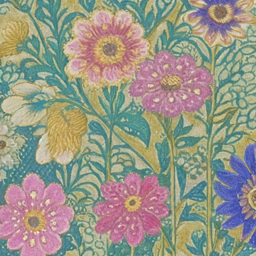}} \\
\hdashline
\textbf{Subject Driven} & Given a set of pictures portraying your neighbor's cat \adjustbox{valign=m}{\includegraphics[width=\xw\linewidth]{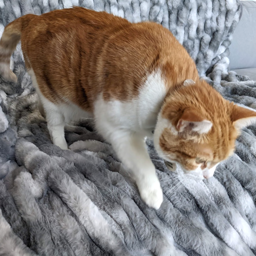}} \adjustbox{valign=m}{\includegraphics[width=\xw\linewidth]{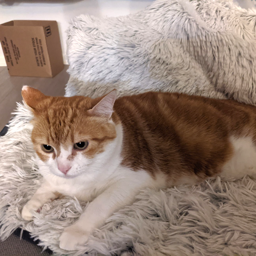}} and \adjustbox{valign=m}{\includegraphics[width=\xw\linewidth]{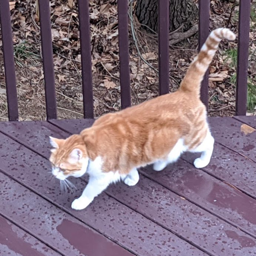}} , create a new image of this cat. & \adjustbox{valign=m}{\includegraphics[width=0.7\linewidth]{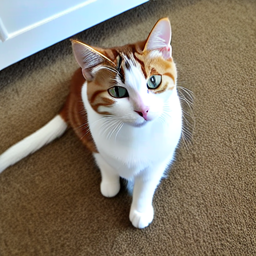}}\\
\bottomrule
\end{tabularx}
\caption{Zero-shot, one-shot, and few-shot image generation examples by \methodname{}. }
\label{tab:task_types_image}
\end{table*}

\newcommand\xa{0.3}
\begin{table*}[!htp]
\centering
\begin{tabularx}{0.999\textwidth}{m{2cm} m{0.62\linewidth} : m{2.5cm}}
\toprule
\textbf{Task Type} & \textbf{Example Prompt} & \textbf{Output}\\
\midrule
\midrule
&\textbf{Zero-Shot Prompting}& \\
\midrule
\textbf{Instruction Editing} & Add echoing to this audio \adjustbox{valign=m}{\includegraphics[width=\xa\linewidth]{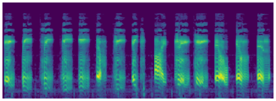}} \textit{(person speaking)}. & \adjustbox{valign=m}{\includegraphics[width=0.99\linewidth]{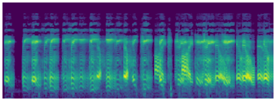}} \textit{(person speaks in echoes)} \\
\midrule
&\textbf{One-Shot/Few-Shot Prompting}&\\
\midrule
\textbf{Exemplar Learning} & We overlaid a new sound to \adjustbox{valign=m}{\includegraphics[width=\xa\linewidth]{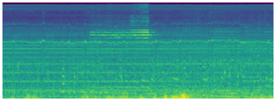}} \textit{(street noise)} and got \adjustbox{valign=m}{\includegraphics[width=\xa\linewidth]{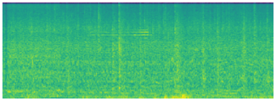}} \textit{(street noise, raining)}. 

Apply this same new sound to \adjustbox{valign=m}{\includegraphics[width=\xa\linewidth]{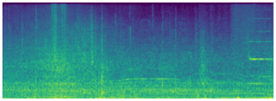}} \textit{(train noise)}. & \adjustbox{valign=m}{\includegraphics[width=0.99\linewidth]{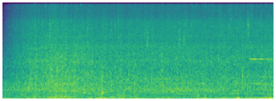}} \textit{(train noise, raining)}\\
\bottomrule
\end{tabularx}
\caption{Audio generation examples of \methodname{}.}
\label{tab:task_types_audio}
\end{table*}

\begin{table*}[!htp]
\centering
\begin{tabularx}{0.999\textwidth}{m{2cm} m{0.62\linewidth}: m{2.8cm}}
\toprule
\textbf{Task Type} & \textbf{Example Prompt} & \textbf{Output}\\
\midrule
\midrule
&\textbf{Zero-Shot Prompting}& \\
\midrule
\textbf{Instruction Editing} & \adjustbox{valign=m}{\includegraphics[width=\xa\linewidth]{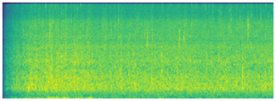}} \textit{(raining)} took place in \adjustbox{valign=m}{\includegraphics[width=\xw\linewidth]{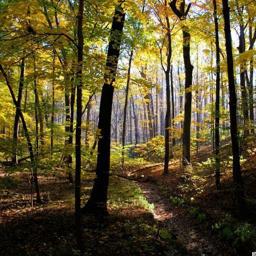}}. & \adjustbox{valign=m}{\includegraphics[width=0.95\linewidth]{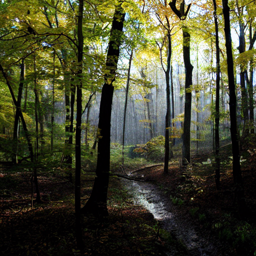}}\\
\hdashline
\textbf{Reasoning} & Given video frames, \adjustbox{valign=m}{\includegraphics[width=\xw\linewidth]{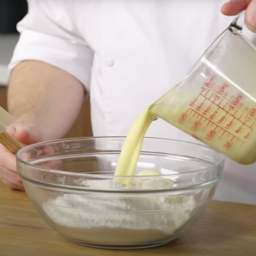}} \adjustbox{valign=m}{\includegraphics[width=\xw\linewidth]{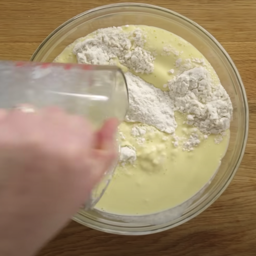}} \adjustbox{valign=m}{\includegraphics[width=\xw\linewidth]{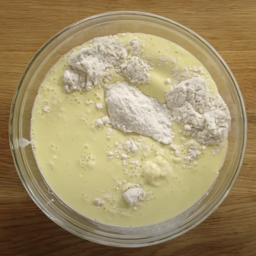}} , 

what will happen next? Generate the sound and image for it. & \adjustbox{valign=m}{\includegraphics[width=0.95\linewidth] {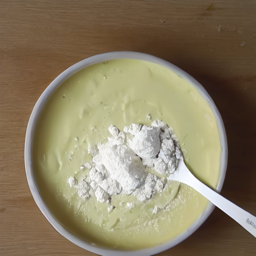}} \adjustbox{valign=m}{\includegraphics[width=0.95\linewidth]{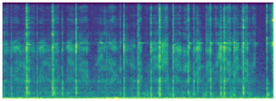}} \textit{(stirring the batter)} \\
\midrule
&\textbf{One-Shot/Few-Shot Prompting}& \\
\midrule
\textbf{Exemplar Learning} & For image \adjustbox{valign=m}{\includegraphics[width=\xw\linewidth]{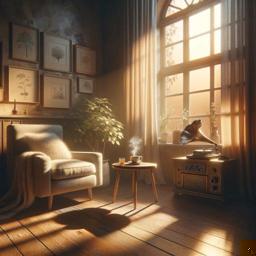}}{ , music} \adjustbox{valign=m}{\includegraphics[width=\xa\linewidth]{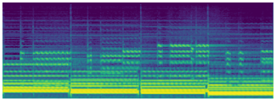}} \textit{(soft jazz music)} {captures its vibe. What's the right music to} \adjustbox{valign=m}{\includegraphics[width=\xw\linewidth]{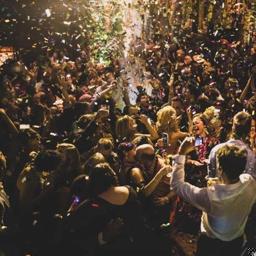}}{ ?} & \adjustbox{valign=m}{\includegraphics[width=0.95\linewidth]{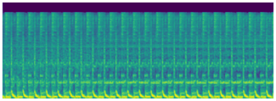}} \textit{(dance music)}\\
\hdashline
\textbf{Subject Driven} & Given a set of pictures portraying your neighbor's cat \adjustbox{valign=m}{\includegraphics[width=\xw\linewidth]{images/table_images/cat_0.png}} \adjustbox{valign=m}{\includegraphics[width=\xw\linewidth]{images/table_images/cat_1.png}} and \adjustbox{valign=m}{\includegraphics[width=\xw\linewidth]{images/table_images/cat_2.png}} , create a video and sound of this cat. & \adjustbox{valign=m}{\includegraphics[width=0.95\linewidth]{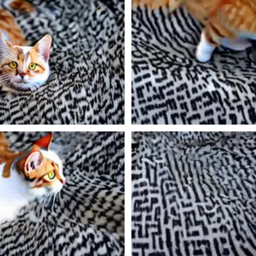}} \adjustbox{valign=m}{\includegraphics[width=0.95\linewidth] {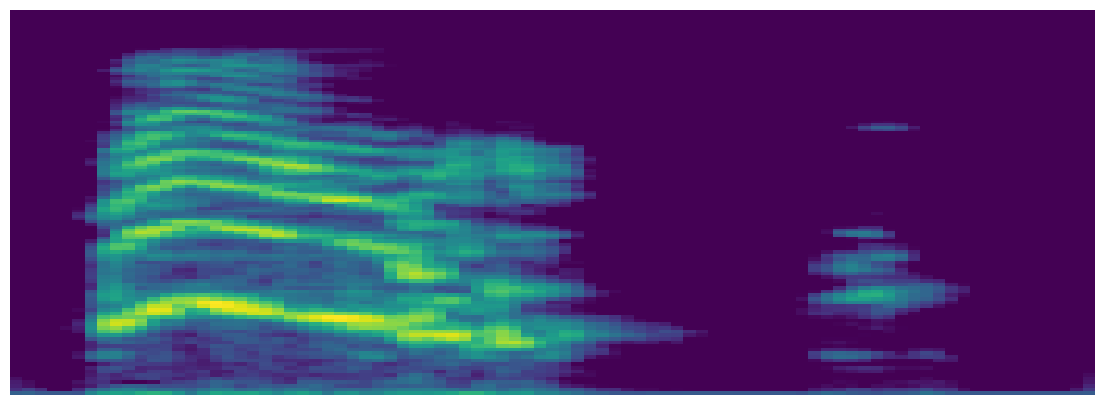}} \textit{(cat meowing)} \\
\bottomrule
\end{tabularx}
\caption{
Example generation with multimodal inputs and outputs. The instructions, image, and audio are interleaved demanding in-context understanding of the inputs. The outputs are either unimodal and multimodal requiring model's synchronized generation abilities. 
} 
\label{tab:task_types_mm}
\end{table*}

\begin{table*}[t]
\centering
\begin{tabular}{lccc}
\toprule Model & DINO $\uparrow$ & CLIP-I $\uparrow$ & CLIP-T $\uparrow$ \\
\midrule \midrule Real Images (Oracle) & 0.774 & 0.885 & - \\
\midrule \multicolumn{4}{c}{ Fine-Tuning } \\
\midrule
DreamBooth~\cite{ruiz2023dreambooth} & 0.668 & 0.803 & 0.305 \\
\midrule\multicolumn{4}{c}{Test Time Tuning Free} \\
\midrule
Re-Imagen~\cite{chen2022re} & 0.600 & 0.740 & 0.270 \\
KoSMOS-G~\cite{pan2023kosmosg} & 0.694 & 0.847 & 0.287 \\
Ours & \textbf{0.703} & \textbf{0.852} & \textbf{0.311} \\
\bottomrule
\end{tabular}
\begin{tabular}{lc}
\toprule
Model & CLIPSIM $\uparrow$ \\
\midrule
\hline \multicolumn{2}{c}{ Diffusion Model Only} \\
\midrule
SDEdit-1/2T~\cite{meng2021sdedit} & 0.134 \\
InstructPix2Pix~\cite{brooks2023instructpix2pix} & 0.151 \\
\midrule
\hline \multicolumn{2}{c}{ Diffusion Model + LLM} \\
\midrule
Ours & 0.147 \\
\bottomrule
\label{tab:image_results}
\end{tabular}
\caption{
Left: Comparisons on DreamBench. Right: Image Editing on MS-COCO.}
\end{table*}

\begin{table*}[t]
\centering
\resizebox{\textwidth}{!}{
\begin{tabular}{cc|ccc|ccc|ccc}
\toprule 
& & \multicolumn{3}{c}{Adding} & \multicolumn{3}{c}{Dropping} & \multicolumn{3}{c}{Replacement} \\
\cmidrule(lr){3-5} \cmidrule(lr){6-8} \cmidrule(lr){9-11}
Model & Text & LSD $(\downarrow)$ & $\mathrm{KL}(\downarrow)$ & $\mathrm{FD}(\downarrow)$ & LSD $(\downarrow)$ & $\mathrm{KL}(\downarrow)$ & $\mathrm{FD}(\downarrow)$ & LSD $(\downarrow)$ & $\mathrm{KL}(\downarrow)$ & $\mathrm{FD}(\downarrow)$ \\
\midrule
SDEdit-3/4T~\cite{meng2021sdedit} & caption & 1.54 & 1.68 & 28.87 & 1.54 & 1.14 & 29.66 & 1.63 & 1.58 & 28.78 \\
SDEdit-1/2T~\cite{meng2021sdedit} & caption & 1.43 & 1.38 & 28.75 & 1.43 & 1.05 & 28.19 & 1.52 & 1.27 & 27.71 \\
SDEdit-1/4T~\cite{meng2021sdedit} & caption & 1.38 & 1.30 & 28.25 & 1.40 & 1.30 & 31.31 & 1.46 & 1.15 & 26.72 \\
AUDIT~\cite{wang2023audit} & instruction & 1.35 & 0.92 & 21.80 & 1.37 & 0.95 & 22.40 & 1.37 & 0.84 & 21.65 \\
Ours & instruction & $\textbf{1.21}$ & $\textbf{0.88}$ & $\textbf{19.72}$ & $\textbf{1.26}$ & $\textbf{0.90}$ & $\textbf{18.06}$ & $\textbf{1.25}$ & $\textbf{0.80}$ & $\textbf{17.32}$ \\
\bottomrule
\end{tabular}}
\caption{Evaluation results of the adding, dropping, and replacement tasks on auditory data. 
}
\label{tab:audio_results}
\end{table*}

\subsection{Dataset Construction}

We construct and employ a variety of datasets to facilitate interleaved and in-context multimodal generation, enriching the capabilities of \methodname{}.

\paragraph{Multimodal In-Context Learning Datasets.} Our approach leverages the strength of multimodal in-context understanding, and to bolster this aspect, we integrate MIMIC-IT~\cite{li2023otter} into our tasks. MIMIC-IT offers an extensive and diverse dataset comprising 2.8 million instruction-response pairs, specifically designed to elevate the performance of Vision-Language Models (VLMs) in real-world scenarios. This augmentation equips VLMs with abilities in perception, reasoning, and planning. Despite its output is text-only, it can help model's in-context understanding of multimodal inputs and overall instruction following. For example in perceptual understanding, given two images with only subtle differences, the instruction is to spot the different. By another example for reasoning, given video frames of football, the instruction is to predict what will happen next.

\paragraph{Multimodal Paired Datasets.}
Paired datasets like image-text are natural multimodal data for cross-modal generation. We use LAION-400M~\cite{schuhmann2021laion400m} that consists of 400 million image-text pairs filtered using CLIP. For audio paired dataset, we use AudioSet~\cite{gemmeke2017audio}. AudioSet offers a comprehensive ontology of 632 audio event classes. It also boasts a collection of 2,084,320 human-labeled 10-second sound clips sourced from YouTube videos. For video paired dataset, we use Webvid~\cite{bain2021frozen}, featuring 10.7 million short video-caption pairs, totaling 52,000 hours, gathered from stock footage websites, showcasing diverse content. We construct two tasks with these datasets, 1) instructing to generate caption given an image or audio, and 2) instructing to generate the image or audio from caption.

\paragraph{Instructional Editing Datasets.}
Instructional Editing is a task structured as an input image, an editing instruction, and the resulting edited image. We use Instructpix2pix~\cite{brooks2023instructpix2pix} for image instructional editing. For audio editing dataset, our approach is built on top of AudioSet~\cite{gemmeke2017audio}. We develop instructional editing versions of it, taking cues from AUDIT~\cite{wang2023audit}. We have developed three versions of this dataset: audio addition (overlay), removal, and replacement, resulting in a dataset three times the size of AudioSet. By overlaying two distinct audio segments, \texttt{a, b}, we obtain a new combined audio \texttt{a+b}. This combined audio can also serve as an input for audio removal \texttt{a+b$\rightarrow$ a}, or removing audio \texttt{b} from audio \texttt{a+b}. Audio replacement is constructed by integrating two different audio segments \texttt{b, c} into the same base audio \texttt{a}, and then we get \texttt{a+b$\rightarrow$ a+c}, or replacing \texttt{b} with \texttt{c} for audio \texttt{a+b} to get \texttt{a+c}.

\paragraph{Constructed In-context Multimodal Generation Datasets.}

To further stimulate the multimodal in-context ability, we construct several in-context datasets for multimodal generation. \textit{InstructPix2Pix} can be extended to interleaved in-context multimodal format,
given its multiple image pairs corresponding to the same editing prompt. Consequently, we define the in-context learning template as: \textbf{Input:} ``\textit{Given the transformation between} \texttt{[image0]} \textit{and} \texttt{[image1]}\textit{, apply the same editing to} \texttt{[image2]}\textit{.}''
\textbf{Target:} \texttt{[image3]}.
This approach can also be adapted for audio editing datasets using the same template structure. 

In addition, we utilize the \textit{Kosmos-G} dataset~\cite{pan2023kosmosg}, constructed using \textit{Open Images V7}~\cite{kuznetsova2020open} with 9M images for image composition. Here, entities from captions are extracted to produce image segmentation for each identity. For instance, for descriptions like `A cat on a couch', we obtain:
\textbf{Input:} \texttt{[image0]} on \texttt{[image1]}
\textbf{Target:} \texttt{[image2]},
where \texttt{[image0]} and \texttt{[image1]} represent the segmented cat and couch images derived from \texttt{[image2]}, respectively. 

\paragraph{Text-Only Datasets Repurposed for Interleaved Multimodal In-Context.}
We propose to employ text-only datasets for enhancing generation with multimodal reasoning. Since the encoder features for all modalities are aligned, replacing text tokens with text encoder features can enhance interleaved multimodal understanding. Concretely, we randomly select phrases or words in the sentence and encode them with text feature encoder to swap out the original text embeddings. This innovative strategy bolsters the model's proficiency in understanding complex, interleaved multimodal scenarios by aligning the encoder features and the original language model text embeddings. We employ instructional dataset alpaca~\cite{taori2023stanford}. For example, we convert `` \textit{A cat typically has a compact, flexible body, covered in soft fur that can come in a variety of colors and patterns.}'' to ``\texttt{[text0]} \textit{typically has} \texttt{[text1]}\textit{, covered in} \texttt{[text2]} \textit{that can come in a variety of colors and patterns.}'' where \texttt{[text0]} \texttt{[text1]} \texttt{[text2]} are respectively text features from ``\textit{a cat}'', ``\textit{a compact, flexible body}'', and ``\textit{soft fur}''.

\subsection{In-Context Instruction Task Types}

\Cref{tab:task_types_image,tab:task_types_audio,tab:task_types_mm} offer a comprehensive overview of the task types utilized in in-context multimodal generation. Each task type presents a unique approach to prompting models to generate or transform in-context multimodal content, including images, audio, and combinations thereof.

\paragraph{Zero-Shot Prompting.}
Zero-shot prompting tasks require the model to reason and generate new content without any prior examples. For instance, in \Cref{tab:task_types_image}, model transforms an image to match Van Gogh's style or compositing two separate images to form a coherent scene exemplifies the model's capacity to understand and apply complex instructions directly. Model also can perform reasoning and predict the next image in a sequence, which is one cube in consistent style. \Cref{tab:task_types_audio} shows adding echoes to an audio. \Cref{tab:task_types_mm} shows visual editing with sound vibe and frame+sound prediction of a video sequence.

\paragraph{One-Shot/Few-Shot Prompting.}
One-shot or few-shot prompting provides the model with one or a few examples to learn from before performing a similar task. This method is evident in tasks where the model adapts a learned concept from one image to another or creates a new piece of artwork by understanding the styles depicted in provided exemplars.

\textbf{Exemplar learning} is a subset of few-shot prompting where the model is explicitly shown an example of the desired output before being asked to apply this learning to a new instance. This technique is particularly useful when trying to generalize a concept from a specific instance to a new, but related, context. In \Cref{tab:task_types_image}, model is shown a set of images with season change and then asked to apply the same to a similar image. In \Cref{tab:task_types_audio}, model is shown a set of audio where the latter one has raining sound on top of it, and then asked to apply the same to a new audio. In \Cref{tab:task_types_mm}, model is shown the audio caption of an image and asked to generate audio for an image with different vibe.

\textbf{Concept learning} involves the model learning from shared concept/attributes of given examples, such as artistic styles or patterns, and then creating new content that exhibits similar concept/attributes. The model's ability to discern and replicate complex patterns indicates a sophisticated understanding of visual styles. In \Cref{tab:task_types_image}, model learns the intricate floral patterns and then draws a new image that reflect the same intricate styles.

\textbf{Subject-driven learning} focus on generating new content based on a set of provided images. This approach tests the model's ability to understand and recreate the subject with variations in pose, lighting, or context, while maintaining the subject's distinct features.
In \Cref{tab:task_types_image}, given several pictures of a specific cat, the model will create a new image of the same cat in new poses. In \Cref{tab:task_types_mm}, given the cat images, the model can create a video+sound of the same cat.

\section{Experiments}
\subsection{Model Setups}
Our implementation is based on Llama2~\cite{touvron2023llama2}, specifically Llama-2-7b-chat-hf. We use ImageBind~\cite{girdhar2023imagebind} which has aligned image, video, audio, text, depth, thermal, and IMU modality encoders. We use ImageBind to encode the image and audio features and project it to the input dimension of the LLM (Llama-2-7b-chat-hf) with a multilayer perceptron (MLP) that consists of a linear mapping, activation, normalization, and one more linear mapping. When the LLM generates the image or audio features, we project them back to ImageBind feature dimension with another MLP. Our image diffusion model is based on StableDiffusion-2.1~\cite{rombach2022high} (stabilityai/stable-diffusion-2-1-unclip~\cite{ramesh2022hierarchical}), AudioLDM2~\cite{liu2023audioldm}, and zeroscope\textunderscore v2\footnote{\href{https://huggingface.co/cerspense/zeroscope_v2_576w}{https://huggingface.co/cerspense/zeroscope\textunderscore v2\textunderscore 576w}}. 

For images or audio that require higher fidelity to the original input, we additionally feed the original image or audio to the diffusion model alongside the generated features by concatenation of the diffusion noise~\cite{rombach2022high,brooks2023instructpix2pix,liu2023audioldm}. This approach is particularly effective in preserving the most perceptual features of the input including instruction editing like adding new content or changing style.

\subsection{Image Generation Evaluation}

\Cref{tab:image_results} shows the evaluation results of subject driven image generation on Dreambench~\cite{ruiz2023dreambooth} and FID scores on MSCOCO. Our method achieves very competitive zero-shot performance, showing our model's generalization to new unseen tasks.

\subsection{Audio Generation Evaluation}
\Cref{tab:audio_results} provides an overview of our evaluation results concerning audio manipulation tasks—namely, adding, dropping, and replacing elements within audio tracks. These results are pivotal in understanding the effectiveness of the proposed methods. It is evident from this table that our approach demonstrates superior performance in comparison to previous methodologies. Notably, it has achieved the lowest scores across all metrics—Log Spectral Distance (LSD), Kullback-Leibler (KL) divergence, and Fréchet Distance (FD)—across all three editing tasks.

%% file: 4_conclusion.tex
\section{Conclusion}

We introduced \methodname{}, a model for multimodal generation with groundbreaking abilities such as modality-interleaved instruction following, in-context generation, user-model interaction through multi-round conversations. \methodname{} is able to processes complex modality-interleaved input and instructions by MLLM, and then autoregressively produce the latent features that is fed to diffusers for multimodal generation. The evaluations show that \methodname{} has exceptional zero-shot and few-shot ability on tasks including style adaptation, subject-driven generation, and editing across modalities. \methodname{} represents a remarkable exploration to build the GPT-like fundamental multimodal system.

%% file: X_suppl.tex
\clearpage
\appendix
\setcounter{page}{1}

\section{Appendix Overview}
\label{sec:supplementary_experiments}

This supplementary section delves deeper into the training methodologies, model details, and experimental setups, as well as the construction of datasets discussed in the main paper. The primary aim is to augment the comprehension of diverse multimodal in-context generation scenarios.

\section{Experiment Setups}
\subsection{Model Setups}
To effectively condition the image diffusion model, we employ negative prompts as cross-attention conditions and utilize MLLM-generated features for embedding guidance~\cite{dhariwal2021diffusion}. The negative prompts used for image generation include: `worst quality, normal quality, low quality, low res, blurry, watermark, logo, banner, extra digits, cropped, jpeg artifacts, signature, username, error, sketch, duplicate, ugly, monochrome, horror, geometry, mutation, disgusting'. For audio, the negative prompts are: distorted, muffled, static noise, background noise, interference, echo, low volume, inaudible, drowned out, screeching, piercing, off-key, out of tune, discordant, interrupted, choppy, glitches, overlapping voices, jumbled, incoherent, repetitive, monotonous, tedious, harsh, grating, abrasive, unbalanced, erratic levels, fluctuating volume, hissing. These negative prompts function as unconditioned input guidance, serving to enhance the quality of both the image and generated features.

\subsection{Training Pipelines}
The training pipeline involves simultaneous text, image, and audio generation via diffusion. To avoid the cumbersome and inefficient aspects of this multitasking approach, especially concerning model I/O, we alternate training phases between text, audio, and image generation. We apply LoRa~\cite{hu2021lora} with a rank of 128 for fine-tuning the model. The fine-tuning process focuses only on the LoRa weights and the projection layers that map modality encoders into the LLM input space, as well as the decoder layer that projects LLM-generated features into the diffusion input space.

\section{Extended Details of Multimodal In-Context}
\label{subsec:extended_multimodal_learning}
This section presents more details on the generation process in multimodal in-context generation datasets.

\subsection{GPT-Generated Prompts} 
We crafted 100 distinct prompt templates for each task type, including instructional editing, multimodal paired datasets, and constructed in-context multimodal generation datasets. For instance, in instructional editing, prompts like `\textit{Given the image} \texttt{[Image0]}, \textit{transform it into Van Gogh style}', or `\textit{Presented with the visual} \texttt{[Image0]}, \textit{convert it into Van Gogh style}' are used. In multimodal paired datasets, we utilize prompts for text-to-image or audio tasks like, `\textit{Generate an image based on the instruction: a cat on a couch}', or `\textit{Produce audio of a person talking}'. For captioning image or audio, examples include `\textit{Generate a caption for this image:} \texttt{[Image0]}' or `\textit{Given} \texttt{[Audio0]}, \textit{produce its description}'. In exemplar learning, a typical prompt is `\textit{Learn the transformation between} \texttt{[Image0]} \textit{and} \texttt{[Image1]}, \textit{and apply it to} \texttt{[Image2]}'. For image composition, prompts like `\textit{Create an image according to the description, combining} \texttt{[Image0]} \textit{with} \texttt{[Image1]}' are used. Each task type includes 100 uniquely generated prompt prototypes, which are sampled uniformly during the data pipeline in training.

\subsection{Addressing the Discrepancy Between Datasets and Practical Applications}
In assembling our research, we have meticulously gathered or integrated a vast array of datasets from various sources. Despite this extensive collection, it is notable that several applications or use-cases remain insufficiently represented within our training datasets. A case in point is visual concept learning, which, although not extensively featured in our training data, is an area where our model excels. Overall, the architecture of our model and the design of our tasks are strategically formulated to leverage the innate in-context capabilities of large language models. This approach is complemented by the use of diverse, in-context, and interleaved datasets, sourced from an array of open-domain materials, thereby enhancing the model's applicability and versatility in addressing a broader spectrum of real-world scenarios.